\documentclass[sigconf, 9pt]{acmart}  %

\usepackage{multirow}
\usepackage{subcaption}
\usepackage{mathtools}
\usepackage{enumitem}
\usepackage{stfloats}
\usepackage{soul}
\usepackage{makecell}
\usepackage{amsmath}
\usepackage{pifont}
\usepackage{algorithm}
\usepackage{algpseudocode}
\usepackage{cleveref}
\usepackage{hyperref}   %
\usepackage{hyperxmp}

\AtBeginDocument{%
  }

\copyrightyear{2025}
\acmYear{2025}
\setcopyright{acmlicensed}\acmConference[TBD '25]{The xxth International Conference on xxx}{TBD, 2025}{location TBD}
\acmBooktitle{The xxth International Conference on xxx, TBD, 2025, TBD}
\acmDOI{xxxx}
\acmISBN{xxxx}

\begin{document}

\title{Budgeted Indirect Adversarial Attack on
Graph-Based Anomaly Detection in Sensor
Networks}

\author{Sanju Xaviar}
\affiliation{%
  \institution{University of Alberta}
  \country{Edmonton, AB, Canada}
}
\email{xaviar@ualberta.ca}

\author{Omid Ardakanian}
\affiliation{%
  \institution{University of Alberta}
  \country{Edmonton, AB, Canada}
}
\email{oardakan@ualberta.ca}

\renewcommand{\shortauthors}{XXX et al.}
\newcommand{\etal}{\textit{et al}.}

\begin{abstract}
Graph Neural Networks (GNNs) have emerged as powerful models for anomaly detection in sensor networks, particularly when analyzing multivariate time series.
In this work, we introduce BETA, a novel indirect evasion attack targeting such GNN-based detectors, where the attacker is constrained to perturb sensor readings from a limited set of nodes, excluding the target sensor, with the goal of either suppressing a true anomaly or triggering a false alarm at the target node. 
BETA uses a graph explanatory model combined with a centrality-based pruning strategy to identify the most influential nodes, subsequently injecting carefully crafted adversarial perturbations into their features.
Extensive experiments on three real-world sensor network datasets show that BETA consistently outperforms baseline attack strategies while operating under realistic constraints, reducing the F1-score of state-of-the-art GNN-based detectors by 36.07 to 50.45\% on average.
\end{abstract}


 
\maketitle

\section{Introduction}
Sensor networks are ubiquitous in critical infrastructure, ranging from water treatment plants and power grids to transportation systems. They continuously generate large volumes of multimodal time-series data, comprising readings from heterogeneous sensors distributed across diverse physical locations. Since these data streams often feed control loops, state estimation algorithms, and event detection models, timely and accurate anomaly detection is paramount for ensuring system reliability and physical safety.
However, detecting anomalies in multimodal time series is inherently challenging.  Sensor nodes exhibit complex spatial and temporal dependencies, while anomalies remain highly sparse~\cite{zhou2025mtsbench}. Traditional statistical and univariate approaches~\cite{paparrizos2022} process sensor channels in isolation, failing to capture these intricate interdependencies and necessitating more sophisticated, structurally-aware detectors.

Graph Neural Networks (GNNs) have emerged as the state-of-the-art paradigm for anomaly detection in sensor networks by modeling dependencies between the sensors~\cite{wu2021graph, kim2022graph, deng2021graph,liu2024multivariate, protogerou2021graph, ma2021comprehensive, jin2024survey}. The sensor network is represented as a graph, where nodes correspond to individual sensors and edges reflect their dependencies. 
This graph-based representation allows anomaly detection to be framed as a node classification problem, where the status of each sensor (normal or anomalous) is inferred considering its own readings and those of its neighbors. Recent advances, such as graph attention networks (GATs), have further enhanced the interpretability of these models by highlighting influential nodes and features~\cite{velivckovic2017graph, wang2019heterogeneous,wang2019kgat}. 
Nevertheless, the reliance of GNNs on message-passing makes them susceptible to \emph{indirect} adversarial perturbations.
If an adversary injects subtle, malicious noise into a sensor's data stream, they can easily manipulate the GNN aggregation and consequently the detection outcome at another node~\cite{tariq2022towards}.

Prior work on adversarial attacks on graph-based machine learning models has primarily focused on unconstrained scenarios or topological manipulation~\cite{zügner2024adversarialattacksgraphneural, dai2018adversarial, sun2022adversarial}. %
For example, the Nettack algorithm \cite{zugner2018adversarial} changes the node classification outcome by injecting small perturbations into the graph structure and node features.
However, sensor networks have the following fundamental differences with other domains where GNNs have been used for classification:
\begin{enumerate}
    \item Critical sensors (e.g., main water valves) are typically safeguarded, preventing direct physical or cyber-intrusion to inject perturbations;
    \item The graph structure is determined by existing physical and functional relationships among sensors, and cannot be altered by an attacker; 
    \item Compromising and maintaining persistent access to sensor nodes is resource-intensive. Thus, an attacker must operate under a strict budget, perturbing only a small subset of peripheral \emph{influencer} nodes.
\end{enumerate}
While budgeted adversarial attacks have been explored in the image domain (such as the one-pixel attack~\cite{su2019one}), transferring this concept to GNNs introduces a complex challenge: how to optimally select a minimal subset of (non-target) influencer nodes whose feature perturbations will maximize the classification error at the target node through the message-passing mechanics of GNNs. Prior work in the area of sensor networks has largely overlooked these realistic constraints, assuming unconstrained budget or access~\cite{tariq2022towards, goodge2021robustness, cao2019adversarial,modas2020toward}.
To address this critical gap, this paper introduces BETA (Budgeted Explainability-guided Targeted Attack), a grey-box evasion attack against GNN-based detectors tailored to the physical constraints of sensor networks.

BETA exploits a GNN explanatory network to identify structural vulnerability in GNN-based anomaly detectors. Specifically, rather than running an intractable combinatorial search over all node subsets, BETA applies GAFExplainer~\cite{Hu2025} to the GNN-based anomaly detector (or its surrogate) to find a candidate subgraph that is most influential in determining the anomaly score of the target sensor. If the subgraph size exceeds the attacker's budget, BETA uses a pruning mechanism to downselect nodes to meet the budget constraint.
Finally, Projected Gradient Descent~(PGD)~\cite{madry2017towards} is executed solely on these selected influencer nodes to inject bounded perturbations.
We evaluate BETA against two state-of-the-art GNN-based anomaly detectors---Graph Deviation Network (GDN)~\cite{deng2021graph} and Topology-aware GDN (TopoGDN)~\cite{liu2024multivariate}---on three real-world testbeds. 
Our contributions are as follows:
\begin{itemize}
    \item We introduce a budgeted adversarial attack tailored to GNN-based anomaly detection in sensor networks, reflecting realistic constraints on attacker capabilities. We formulate this indirect adversarial attack as a joint node selection and  perturbation optimization problem.

    \item Since this problem is intractable due to its combinatorial nature,
    we propose a novel methodology that relies on post-hoc GNN explanations to reduce the search space for the attacker, allowing it to efficiently generate adversarial perturbations using PGD. 
    To ensure that the attack meets the strict budget constraint, we refine the candidate influencer nodes according to their eigenvector centrality;
    this ensures that perturbations diffuse as widely as possible through the GNN,
    increasing the likelihood of affecting the target node's representation and detection outcome.
    
    \item Through extensive evaluation on three datasets, we show that BETA consistently outperforms baseline attacks, and causes the detector's F1-score to decrease by 36.07 to 50.45\%, on average, compared to its performance on the original data. These results expose a critical vulnerability: protecting the target sensor is entirely ineffective if the surrounding graph neighborhood is left unsecured.

\end{itemize}

\begin{figure}[t!]
\centering
\includegraphics[width=\linewidth]{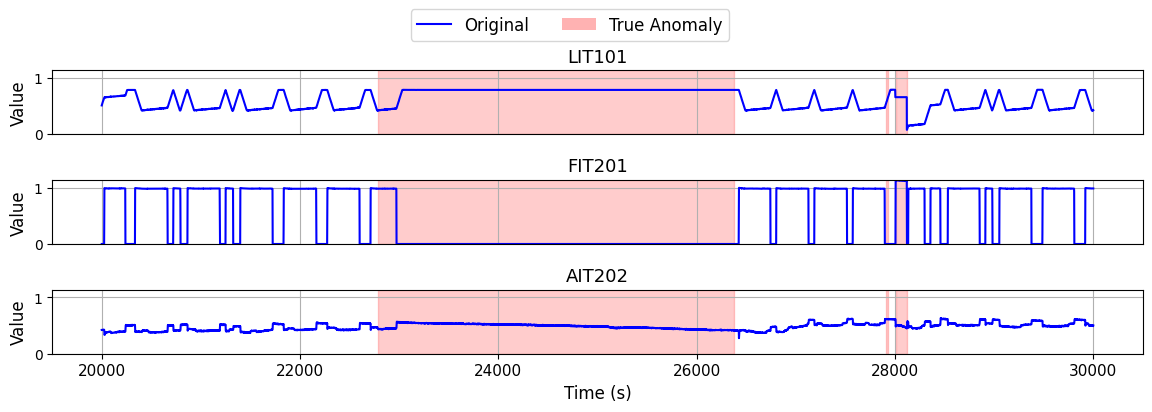}
\vspace{-5mm}
\caption{Visualization of the readings of three sensors in the SWaT dataset, with time intervals that contain anomalies highlighted in red. 
}
\label{fig:SWATAnomaly}
\end{figure}

\begin{figure}[t!]
\centering
\includegraphics[width=\linewidth]{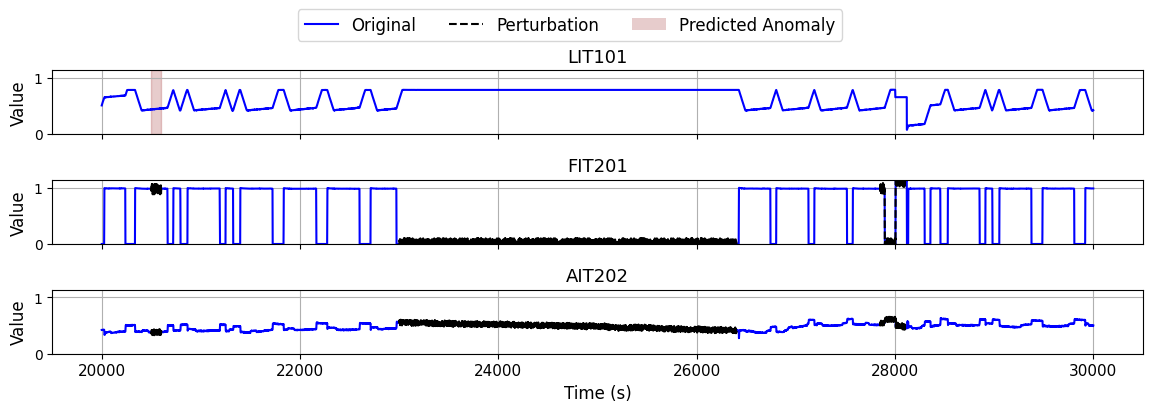}
\vspace{-5mm}
\caption{Visualization of perturbed sensor readings in the SWaT dataset. FIT201 and AIT202 are selected as influencer nodes, LIT101 is selected as the target node.
}
\label{fig:SWATAnomaly_afterattack}
\end{figure}

Figure~\ref{fig:SWATAnomaly} shows the original readings of three representative sensors, \textit{i.e.} LIT101 (tank level), FIT201 (flow) and AIT202 (chemical analyzer), in the SWaT dataset (introduced in Section~\ref{sec:datasets}). The shaded regions are intervals that contained true anomalies.
Figure~\ref{fig:SWATAnomaly_afterattack} illustrates how BETA adds imperceptible perturbations to readings of FIT201 and AIT202 so as to change the anomaly detection outcome for sensor LIT101, which is assumed to be the target node.
Despite not being able to perturb readings of LIT101 directly, the GNN-based anomaly detector falsely detects anomalies and suppresses true anomalies that existed in the dataset.

\noindent

\section{Related Work}

Adversarial attacks manipulate machine learning models by introducing carefully crafted perturbations to input data. The goal is to cause misclassification of data that belong to either a specific class or any class.
Adversarial attacks can also be classified based on the level of knowledge that the attacker has about the machine learning model into three categories: 
white-box, grey-box, and black-box attacks~\cite{chen2020survey}. In a white-box attack, the attacker has full access to the model architecture and parameters, prediction results, and even gradient information during training. 
With this knowledge, the attacker can generate highly effective adversarial samples. 
In a grey-box attack, the attacker has only partial knowledge of the model, such as limited access to the model architecture or a subset of model parameters. 
In a black-box attack, the attacker has no information about the model; it has only query access to the model~\cite{sun2022adversarial}, \textit{i.e.} it can observe the model's output for a given input.
This query access is used to train a surrogate model to perform the attack.

Adversarial attacks can be performed at training time (poisoning attacks) and at test time (evasion attacks). 
Evasion attacks, such as 
FGSM~\cite{goodfellow2014explaining} and
PGD~\cite{madry2017towards}, 
mislead a trained model by modifying certain features in the input data.
Graph-based evasion attacks, such as GradArgmax~\cite{dai2018adversarial}, may also alter links in the underlying graph based on gradient information from a surrogate model to deceive the classifier. 
We focus on evasion attacks that add perturbations to sensor readings only (no changes to graph structure) as these attacks are more applicable to sensor networks.
In particular, we propose a constrained PGD-based evasion attack tailored for anomaly detection in sensor networks, under the grey-box assumption, i.e. the attacker knows the graph structure of the GNN model used for anomaly detection.

\subsubsection{Evasion Attacks on Graph Neural Networks}

The most notable example of adversarial attacks against GNNs is Nettack \cite{zugner2018adversarial}, which aims to misclassify a set of target nodes. 
This attack is grey-box because it assumes access to the graph structure, node features, and the overall architecture of the target model, but not its parameters.
The authors study poisoning or causative attacks, in addition to evasion attacks.
Their approach involves generating adversarial perturbations that target both node features and the graph structure, accounting for dependencies between instances. 
In~\cite{zhang2021projective}, an evasion attack, called projective ranking, is designed to generate transferable adversarial perturbations for graph neural networks. It 
incorporates mutual information to assess the long-term impact of perturbations. This approach enables transferable, structure-based (rather than feature-based) attacks across varying target nodes and perturbation budgets.

\begin{table*}[ht!]
\centering
\small
\caption{Evasion attacks against multivariate time series anomaly detection (D indicates direct attack and I indicates indirect attack via influencer nodes).}
\label{tab:method_feature_comparison}
\begin{tabular}{l l l c c}
\toprule
\textbf{References} & \textbf{Attack} & \textbf{Setting} & \textbf{D/I} & \textbf{Budgeted} \\
\midrule
Tariq et al.~\cite{tariq2022towards} & FGSM, PGD & White-box, Black-box & D & \ding{55} \\
Goodge et al.~\cite{goodge2021robustness} & FGSM & White-box & D & \ding{55} \\
Jia et al.~\cite{jia2021adversarial} & FGSM & White-box & D & \ding{55} \\
Srinidhi et al.~\cite{madabhushi2025mitigating} & FGSM, BIM, PGD & White-box & D & \ding{55} \\
\textit{BETA (Our approach)} & Constrained-PGD & Grey-box & I & \ding{51} \\
\bottomrule
\end{tabular}
\end{table*}

\subsubsection{Evasion Attacks on Anomaly Detection Models}
Table~\ref{tab:method_feature_comparison} summarizes the prior work on evasion attacks targeting multivariate time series anomaly detection models. Most previous work focuses on gradient-based attacks, such as FGSM and PGD, typically in a white-box setting where the model architecture and parameters are fully accessible. 
Moreover, all these attacks are direct, meaning that the features of the target node are directly perturbed, 
overlooking indirect manipulation and budgeted settings. 
In contrast, our approach proposes a constrained-PGD attack in a grey-box setting, capable of performing indirect perturbations under a strict budget, making it more practical and targeted for real-world multivariate anomaly detection scenarios. 
It is worth noting that PGD has been found to consistently outperform FGSM and other gradient-based evasion attacks in the multivariate case~\cite{tariq2022towards, madabhushi2025mitigating}, 
motivating us to build BETA on top of PGD.

\section{Preliminaries}\label{sec:prelim}
We provide an overview of two representative GNN-based models for multivariate time series anomaly detection that utilize graph structure learning and attention, namely the Graph Deviation Network (GDN)~\cite{deng2021graph} and its extension, TopoGDN~\cite{liu2024multivariate}.
These two models are used to evaluate BETA.

GDN~\cite{deng2021graph} detects anomalies by leveraging a GNN to capture both dependencies within the sensor data.
It consists of two main components: a time series forecasting model and a threshold-based anomaly detection mechanism. The forecasting module uses GAT to predict future sensor readings based on historical data and learned relationships. The anomaly detection mechanism then evaluates discrepancies between predicted and actual readings, assigning anomaly scores accordingly. TopoGDN~\cite{liu2024multivariate} extends GDN by incorporating a multi-scale temporal convolution module that captures fine-grained short and long-term temporal dependencies in sensor readings and a topological graph attention and pooling module that enhances structural representations by integrating higher-order topological features derived via persistent homology, while preserving the core architecture of GDN, described in Sections~\ref{sec:input_representation} to~\ref{sec:anomaly_detection}. The architectural enhancements specific to TopoGDN are detailed in Sections~\ref{sec:multi_scale_temporal_conv} and~\ref{sec:topological_attention_pooling}.

\subsection{Graph Structure Learning}
\label{sec:input_representation}
Consider a multivariate time series segment \(\mathbf{X}^t = [ \mathbf{x}^t_1, \dots, \mathbf{x}^t_N ]^\top\),
where \( N \) is the total number of sensors and 
each \( \mathbf{x}^t_i \) collects historical data emitted by sensor \( i \) in a time window of length \( w \), preceding time \( t \):
\begin{equation} 
\mathbf{x}^t_i = [x^{(t-w)}_i,\cdots,x^{(t-1)}_i]\in \mathbb{R}^w
\end{equation}
The GDN model learns relationships among sensors using sensor embedding vectors \( \mathbf{v}_i \in \mathbb{R}^d \) for each sensor \( i \), which are initialized randomly and trained jointly with the rest of the model parameters via backpropagation. 
These embeddings define a similarity metric given by:
\begin{equation}
    g_{ji} = \frac{\mathbf{v}_i^T \mathbf{v}_j}{\|\mathbf{v}_i\| \|\mathbf{v}_j\|} \mathbb{I}(j \in \mathcal{C}_i)
\end{equation}
where \( \mathcal{C}_i \) represents a set of candidate connections for sensor \( i \), $\|.\|$ denotes the Euclidean norm, and the indicator function \( \mathbb{I} \) returns 1 when node $j$ belongs to \( \mathcal{C}_i \) and 0 otherwise. 
A graph adjacency matrix \( \mathbf{A} \) is then constructed by connecting each node to its \( M \) most similar neighbors, with respect to \( g_{ji} \), where \( M \) defines the maximum number of edges a node can have, which is a hyperparameter.

\subsection{GAT for Feature Extraction}
\label{sec:gat_feature_extraction}
The GDN model employs a GAT layer to dynamically aggregate information from a node's neighbors. For each node \( i \), the aggregated representation \( \mathbf{r}_i^{t} \) is computed as:
\begin{equation}\label{eq:agg-rep}
    \mathbf{r}_i^{t} = \text{ReLU} \left(\sum_{j: \mathbf{A}_{ji} > 0} \beta_{ij} \mathbf{W} \mathbf{x}_j^{t} \right)
\end{equation}
where \( \mathbf{x}_j^{t} \) represents the feature vector of node \( j \) at time \( t \), and \( \mathbf{W}\in\mathbb{R}^{d\times w} \) is a learnable weight matrix. 
The attention coefficients \( \beta_{ij} \) are computed as:
\begin{equation}
    \beta_{ij} = \frac{\exp(\rho_{ij})}{\sum_{k: \mathbf{A}_{ki} > 0} \exp(\rho_{ik})}
\end{equation}
where
\begin{equation}
    \rho_{ij} = \text{LeakyReLU} \left( \omega^\top \left( (\mathbf{v}_i \oplus \mathbf{W} \mathbf{x}_i^{t}) \oplus (\mathbf{v}_j \oplus \mathbf{W} \mathbf{x}_j^{t}) \right) \right)
\end{equation}
with a learnable parameter vector \( \omega \), and $\oplus$ for concatenation. 

\subsection{Anomaly Detection Mechanism}
\label{sec:anomaly_detection}
Next, \textit{anomaly scores} are computed based on forecasting errors. The prediction \( \hat{x}^{(t)} \) is obtained by passing the aggregated node representations to a number of stacked fully-connected layers denoted as \(h_\theta\):
\begin{equation}
    \hat{\mathbf{x}}^{(t)} = h_\theta \left( \mathbf{v}_1 \odot \mathbf{r}_1^{t}, \dots, \mathbf{v}_N \odot \mathbf{r}_N^{t} \right),
\end{equation}
where \( \odot \) denotes element-wise multiplication. 
For each sensor \( i \), the prediction error is computed as:
\begin{equation}
    \xi_{i,t} = \left| x_i^{(t)} - \hat{x}_i^{(t)} \right|,
\end{equation}
and is standardized %
to obtain the anomaly score:
\begin{equation}
    \tilde{\xi}_{i,t} = \frac{\xi_{i,t} - \text{Median}}{\text{IQR}}.
\end{equation}
In the above equation, \(\text{Median}\) and \(\text{IQR}\) are respectively the median and interquartile range of \( \xi_{i,t} \) across time.
To compute the overall anomaly detection outcome at time step $t$, the prediction 
errors across sensors are aggregated using the max function, then compared against a threshold:
\begin{equation}
    y^{(t)} = \mathbb{I}\left( \max_i(\tilde{\xi}_{i,t}) > \lambda \right),
\end{equation}
where $\mathbb{I}$ is an indicator function.
Since we focus on \textit{node-level} anomaly detection, we omit the hardmax aggregation in the last step to obtain the anomaly detection outcome for each node $i$ at time $t$:
\begin{equation}\label{eq:thresholding}
y^{(t)}_i = \mathbb{I}\!\left( \tilde{\xi}_{i,t} > \lambda \right).
\end{equation}

\subsection{Multi-Scale Temporal Convolution}
\label{sec:multi_scale_temporal_conv}
TopoGDN captures multi-scale temporal patterns using 1D convolutions with varying receptive fields. 
It applies \( H \) convolution filters, each with kernel size \( w_h \) ($1\leq h \leq H$), to extract temporal features at different scales. 
The output of the \( h \)-th filter at time step \( t \) is computed as:
\begin{equation}
    z^{(h)}_i[t] = \sum_{l=0}^{w_h - 1} x^{(t + l)}_i \cdot f_h[l],
\end{equation}
where $f_h[{i}]$ denotes the learnable weights of the \( h \)-th convolutional filter at position \(i\).
The aggregated output at time step \( t \) is then given by:
\begin{equation}
    z_{i}[t] = \text{Pool}\left(z^{(1)}_i[t], z^{(2)}_i[t], \dots, z^{(H)}_i[t]\right),
\end{equation}
where \( \text{Pool}(\cdot) \) denotes a pooling operation (e.g., average or max), which merges the outputs of all convolution scales into a unified temporal representation for node \( i \). 

The resulting $\mathbf{z}_{i}^t=\big[z_{i}[t-w],\cdots,z_{i}[t-1]\big]\in\mathbb{R}^w$ is then used instead of \( \mathbf{x}_i^{t} \) to compute the initial and aggregated node representations following Eq.~\ref{eq:agg-rep}.

\subsection{Topological Graph Attention and Pooling}
\label{sec:topological_attention_pooling}
TopoGDN builds upon the GAT mechanism used in GDN (Section~\ref{sec:gat_feature_extraction}) by introducing two enhancements. 
First, it replaces the standard dot-product attention with a feed-forward network. 
Second, after computing node embeddings using attention scores, it applies topological pooling based on persistent homology~\cite{liu2024multivariate}. Specifically, it extracts high‑order topological features from a sequence of graph filtrations and converts them into persistence diagrams, which are then vectorized and fused with the final embeddings to enrich node representations.
The resulting node representations are used to predict future values and compute anomaly scores.

\section{Problem formulation}
We consider a sensor network that consists of $N$ sensors installed at different locations, each generating real values at regular intervals. 
We model the dependency between these sensors using a weighted graph \( G = (\mathcal{V}, \mathcal{E}) \), where \( \mathcal{V} \) denotes the set of sensor nodes, and \( \mathcal{E} \) denotes the set of weighted edges with weight $e_{ij}$ representing the degree of dependency between sensor $i$ and sensor $j$, which is time-varying.
The learned adjacency matrix \( \mathbf{A} \in \{0, 1\}^{N \times N} \) captures the structure of this graph, which does not change over time.

Given a multivariate time series segment \( \mathbf{X}^t \in\mathbb{R}^{N \times w} \), with \( w \) being the size of the window (spanning time steps \( t-w \) to \( t-1 \)) used to segment the data, 
the anomaly detection models described in Section~\ref{sec:prelim} decide whether to mark the reading of each sensor at time \( t \) as an anomaly. 
To this end, they learn a function \( f_\theta(\mathbf{X}^t; \mathbf{A}) \) to predict the reading of each sensor at \( t \), given the learned adjacency matrix \( \mathbf{A} \). 
The training data contains only normal time series segments, and the forecasting model is trained using the MSE loss.

At test time, the anomaly detection model receives a multivariate time series segment and decides whether the data emitted by each sensor in the next time step is anomalous by comparing the normalized prediction error against a threshold \( \lambda \). 
As a result of this comparison, a vector \( \{0, 1\}^N \) is formed, where each element of this vector indicates whether the reading of the respective sensor at \( t \) is an anomaly.
Since the test data contains real anomalies, BETA aims to degrade the performance of the anomaly detection model by manipulating the anomaly score for the target node $u\in \mathcal{V}$. 

\begin{figure}[t!]
\centering
\includegraphics[width=\linewidth]{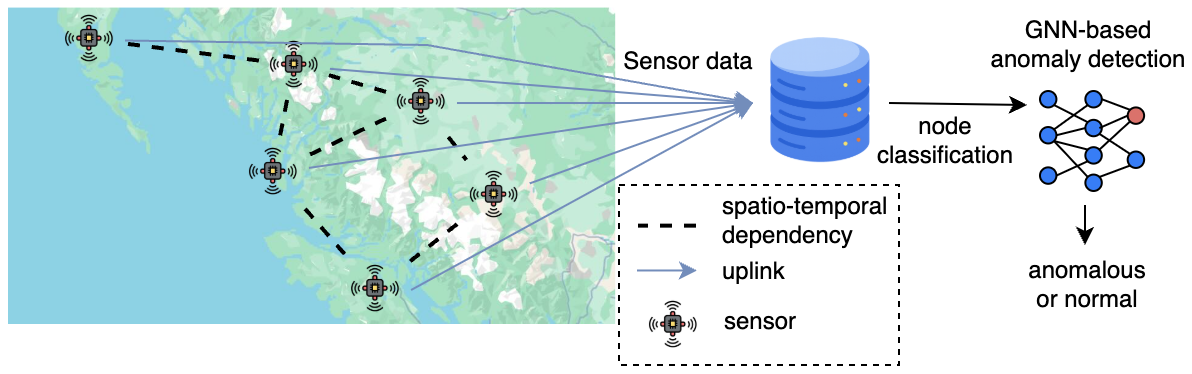}
\vspace{-5mm}
\caption{Illustration of a scenario where sensors installed at different locations collect time series data and transmit them to a central node where a GNN-based anomaly detection model is deployed. An attacker may compromise a small number of strategically chosen influencer nodes to add perturbations so as to mislead this model.}
\label{fig:GNN_Anomalydetection}
\end{figure}

\textit{Assumptions.} We assume that (1) the GNN-based anomaly detection model is deployed on a central/sink node as sensor nodes are resource-constrained, lacking the computational power to run the detection model; the sink node receives data from all sensors in the sensor network, including the target sensor, as depicted in Figure~\ref{fig:GNN_Anomalydetection}; 
(2) the target sensor, which is a critical sensor, is safeguarded and cannot be compromised unlike other sensors; 
(3) sensor nodes transmit their data to the sink node in segments, i.e., they buffer their readings locally and once a full segment is ready, it is sent out; 
this is necessary to avoid draining the 
sensor batteries and optimize network bandwidth;
(4) these transmissions occur asynchronously due to variable network latency and clock drift. The sink node resolves this by utilizing a temporal alignment buffer to assemble data before feeding them to the GNN.

\subsection{Threat model}

The attack considered in this work is a feature-based, indirect evasion attack against GNNs with known graph structure.
The goal of the attacker is to suppress a true anomaly or trigger a false alarm at the target node.
To this end, the attacker can add perturbation to the input features of the deployed anomaly detection model, i.e. the \textit{victim model}, by compromising a subset of the sensors in the network $\bar{\mathcal{V}}\subset \mathcal{V}$, referred to as \textit{influencer nodes}, which are distinct from the target sensor. 
The cardinality of $\bar{\mathcal{V}}$ is constrained by the attacker's budget denoted as $B$.
The attacker is capable of intercepting the communication between each sensor node and the sink node.
Since segments are transmitted asynchronously by the sensors, the attacker can slightly delay segment transmission by the compromised (influencer) nodes to create an opportunity to intercept the segments transmitted by the other nodes before transmitting the influencer nodes' segments. 
This allows the attacker to obtain the full feature matrix \( \mathbf{X}^t \) and use it to compute the amount of perturbation that should be added (Algorithm~\ref{alg:beta}). 

The attacker operates under a grey-box assumption, with full access to the learned graph structure \( \mathbf{A} \), but no access to victim model parameters, layers, activation functions, training data, or gradients.\footnote{\textcolor{black}{This assumption is plausible in real-world sensor network deployments where the graph structure is either disclosed for regulatory or operational reasons, or can be easily inferred from historical sensor data.
It is also strictly weaker than the white-box assumption made in prior work (see Table~\ref{tab:method_feature_comparison}), as it does not require access to learned parameters.}}
It also has query access to the victim model \( f_\theta \), enabling it to train a \textit{surrogate model} denoted as \( f_{\zeta} \) to approximate the victim model's behavior.

\subsection{Node Selection and Perturbation Generation}
To perform the evasion attack, the attacker needs to select the set of influencer nodes, then generate an adversarial feature matrix \( \mathbf{X}'^t = \mathbf{X}^t + \Delta \) by computing a perturbation matrix $\Delta$ subject to the budget constraint and the bounded perturbation (stealthiness) constraint. 
In BETA, the surrogate model guides the selection of the influencer nodes and the computation of \( \Delta \).
Let $\mathcal{L}\big(f_{\zeta}(\mathbf{X}^t; \mathbf{A}),y^t\big)$ be the loss of the surrogate model with $y^t$ representing the true anomaly label for the target node. The joint influencer node selection and perturbation optimization problem can be written as follows:
\begin{align*}
    \max_{\bar{\mathcal{V}}\subseteq \mathcal{V}\setminus\{u\},|\bar{\mathcal{V}}|\leq B}~ &\max_{\Delta} &&\mathcal{L}\big(f_{\zeta}(\mathbf{X}^t+\Delta; \mathbf{A}),y^t\big)\\
    &\text{subject to}&&\\
    & &&\|\Delta_{i}\|_0 = 0 ~~~~\forall i\ne \bar{\mathcal{V}}\\
    & &&\| vec(\Delta) \|_\infty \leq \epsilon
\end{align*}
Here, $u$ is the target node, $vec(.)$ is the vectorization operator, and $\Delta_{i}$ represents the i\emph{th} row of matrix $\Delta$.
While the inner optimization problem can be solved by linearizing the loss around $\mathbf{X}^t$, as in FGSM and PGD attacks, the outer problem is combinatorial and intractable.
Therefore, instead of resorting to combinatorial search, we utilize a graph explanatory network to identify a candidate set $\bar{\mathcal{V}}$ and enforce the budget constraint by refining $\bar{\mathcal{V}}$ according to eigenvector centrality.

\section{Methodology}
\begin{figure}[t!]
\centering
\includegraphics[width=\linewidth]{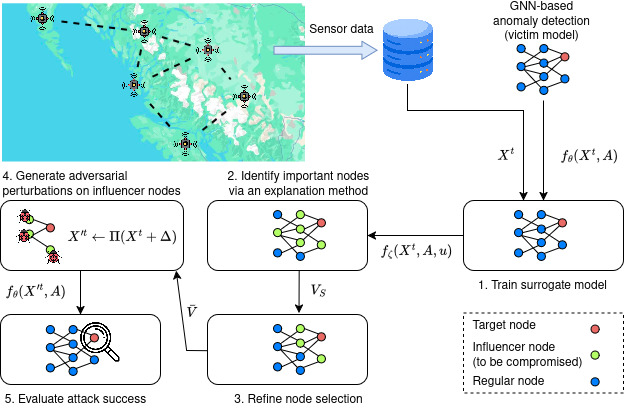}
\vspace{-5mm}
\caption{Illustration of BETA attack pipeline}
\label{fig:Proposedframework}
\end{figure}

This section outlines the graph explanation module employed for selecting influencer nodes (Section~\ref{sec:gafexplainer}) and the topology-based heuristic used to prune the set of candidate nodes (Section~\ref{sec:centrality}). 
Figure~\ref{fig:Proposedframework} describes different steps of BETA, from training a surrogate model, to identifying candidate influential nodes and pruning it, to performing the evasion attack by adding carefully crafted noise to features of these nodes.

Since the attacker does not have knowledge of the victim model, it has to train a surrogate model to approximate the anomaly detection outcome of the victim model for the target node. 
Using query access to the victim model, 
the attacker constructs a dataset for training the surrogate model consisting of multiple pairs of historical measurements and the label for the target node, i.e., $\{(\mathbf{X}^t, y_{u}^{(t)})\}_{t=1,\cdots,l}$. 
Hence, the surrogate model integrates a forecasting module with a threshold based anomaly detection module (see Eq.~\ref{eq:thresholding}), which outputs a binary anomaly label for the target node (0 = normal, 1 = anomalous).
The surrogate model is trained using the cross-entropy loss.

Algorithm~\ref{alg:beta} presents the steps of BETA.
The attacker first chooses the influencer nodes given the strict budget $B$, the adjacency matrix $\mathbf{A}$, and the surrogate model $f_{\zeta}$. 
We outline this process in the following subsections. 
After obtaining the set of influencer nodes, denoted as $\bar{\mathcal{V}}$, and compromising these sensors, the attacker uses PGD to iteratively perturb the node features of the compromised sensors.
Specifically, we initialize PGD by uniformly sampling an initial feature matrix from an $\ell_\infty$-ball around the original input and masking the elements in that matrix that correspond to the non-influencer nodes. This gives $\mathbf{X}_{0}$ defined below: 
\begin{align}
\mathbf{X}_{0} &= \tilde{\mathbf{X}} \odot \mathbf{S}_{\bar{\mathcal{V}}},~~~~\tilde{\mathbf{X}} \sim \mathcal{B}_\infty(\mathbf{X}^t, \epsilon),\\ \nonumber
\, \mathcal{B}_\infty(\mathbf{X}^t, \epsilon) &= \{ \mathbf{X} : \|vec(\mathbf{X} - \mathbf{X}^t)\|_\infty \leq \epsilon \},
\end{align}
where $\odot$ is Hadamard product, and $\mathbf{S}_{\bar{\mathcal{V}}}$ denotes a binary selector matrix with rows corresponding to the selected influencer nodes are vectors of 1 and other rows are vectors of 0.

At each iteration $k$, the features are updated in the direction of the gradient of the loss function $L$, 
which is the cross-entropy loss between the prediction of the surrogate model $f_{\zeta}$ for the target node $u$ and the true anomaly label of the target node, which is denoted as
$\nabla_{\mathbf{X}} L(f_{\zeta}(\mathbf{X}_k;\mathbf{A}), y)$
where $\mathbf{X}_k$ denotes the feature matrix at iteration $k$, and $\nabla_\mathbf{X} L(\cdot)$ computes the gradient of the loss with respect to the input. 
Finally, the influencer nodes' features are updated according to the PGD update rule:
\begin{equation}\label{eq:pgd}
\mathbf{X}_{k+1} = \Pi\left( \mathbf{X}_k + \alpha {\cdot} \text{sign}(\nabla_{\mathbf{X}} L(f_{\zeta}(\mathbf{X}_k,\mathbf{A}), y)) \right)\odot \mathbf{S}_{\bar{\mathcal{V}}},
\end{equation}
where
$\alpha$ is the step size, $\text{sign}(\cdot)$ denotes the element-wise sign function, and $\Pi(\cdot)$ is the projection operator that ensures the perturbed features remain within the $\ell_\infty$-ball around the original input. This iterative update continues until the maximum number of iterations is reached. Then, the final adversarial feature matrix $\mathbf{X}'^t$ is returned.
The goal of this update is to maximize the classification loss, i.e., make the model misclassify the target node, while ensuring the perturbations remain small and stealthy.

\begin{algorithm}[!t]
\caption{BETA}
\label{alg:beta}
\begin{algorithmic}[1]

\State \textbf{Input:} 
Adjacency matrix $\mathbf{A} \in \mathbb{R}^{N \times N}$, 
Original feature matrix $\mathbf{X}^t \in \mathbb{R}^{N \times w}$, 
Surrogate model $f_{\zeta}$, 
Target node $u$, 
Ground truth anomaly label of target node $y$, 
Number of iterations $K$, 
PGD step size $\alpha$.

\State \textbf{Output:} Perturbed feature matrix $X'^t$ %

\State $\mathcal{V}_S \gets \texttt{GAFExplainer}( \mathbf{X}^t,\mathbf{A}, u, E)$ \Comment{Select nodes connected by $E$ most important edges}
\State $\bar{\mathcal{V}} \gets \texttt{SelectTopNodes}(\mathcal{V}_S, B)$ \Comment{Using centrality measures}
\State Initialize 
$\mathbf{X}_0 = \tilde{\mathbf{X}} \odot \mathbf{S}_{\bar{\mathcal{V}}}$ where
$\tilde{\mathbf{X}} \sim \mathcal{B}_\infty(\mathbf{X}^t, \epsilon)
$

\While{$k < K$}

    \State $\resizebox{.94\linewidth}{!}{$\mathbf{X}_{k+1}=\Pi\!\left(\mathbf{X}_k+\alpha\,\mathrm{sign}\!\big(\nabla_{\!\mathbf{X}}L(f_{\zeta}(\mathbf{X}_k,\mathbf{A})_u,y)\big)\right)\!\odot\mathbf{S}_{\bar{\mathcal{V}}}$}$

\EndWhile

\State \textbf{return} $\textbf{X}'^t \gets \textbf{X}_K$

\end{algorithmic}
\end{algorithm}

\subsection{Identifying Candidate Influential Nodes}
\label{sec:gafexplainer}
To identify influential nodes for adversarial perturbation, we apply \textit{GAFExplainer}~\cite{Hu2025} to the surrogate model $f_{\zeta}$, which identifies an important subgraph of this GNN by taking into account its topology and node attributes.
The surrogate model produces the prediction $y_{\mathcal{O},u}$ for the target node $u$, together with intermediate node embeddings from each of its $L$ layers.
These embeddings are used by GAFExplainer to trace the regions of the graph that most influence the target node's output. GAFExplainer first uses a \textit{node attribute augmentation} module, 
which uses an attention mechanism to aggregate embeddings from the neighbors of each node, producing enhanced node features that reflect both temporal and structural context. These augmented features are combined with layer-wise GNN embeddings using a \textit{fusion embedding} module. Edge embeddings are then passed through a learnable \textit{weight generator}, 
followed by a \textit{mask fuser}, 
which uses a binary concrete distribution to construct a pseudo-discrete mask over the edges. 
The resulting mask extracts a candidate subgraph, which is re-evaluated by the surrogate GNN to produce a subgraph prediction \( y_{\mathcal{S},u} \). 
The explanatory network is trained by minimizing the difference between \( y_{\mathcal{S},u} \) and $y_{\mathcal{O},u}$, allowing it to identify the most influential nodes for classifying the target node. 

This process produces a subset of nodes that might exceed the attack budget and include the target node itself; in that case we remove it from the set.
We introduce a hyperparameter $E$ to control the number of edges that can be kept in the explanation subgraph. 
Once these top-ranked edges are selected, their incident nodes are automatically included in the subgraph and are treated as the candidate influential nodes.
In practice, we choose $E = B - 1$, because if the identified subgraph is connected, we will get either less than $B$ nodes or exactly $B$ nodes (if it is a tree). 
However, the identified subgraph may have multiple connected components, resulting in more than $B$ nodes.
As choosing a certain value of $E$ does not guarantee that exactly $B$ nodes will be included in the subgraph, the following step is required to enforce the budget $B$.

\subsection{Pruning Influential Nodes}
\label{sec:centrality}
Once the set of candidate influencer nodes, denoted as $\mathcal{V}_S$, is produced by GAFExplainer, 
we further prune it by choosing the top-$B$ influential nodes according to a widely used centrality measure, namely eigenvector centrality~\cite{newman2010networks}.
Eigenvector centrality assesses a node's importance by considering the centrality of its neighbors. In other words, a node is considered influential if it is connected to other influential nodes. 
Let vector $c$ collect the eigenvector centrality of all nodes in graph $G$. 
We have $\mathbf{A}c=\lambda_{max}c$ where $\lambda_{max}$ is the largest eigenvalue of $\mathbf{A}$. 
We rank the nodes in $\mathcal{V}_S$ according to their eigenvector centrality and select the top $B$ nodes among them. This yields the set of influential nodes, denoted as $\bar{\mathcal{V}}$. The features of these nodes are then perturbed by PGD.

We note that the proposed pruning strategy ensures that the nodes selected for perturbation maximize both the influence on the target node classification outcome and the multi-hop diffusion of the injected perturbations.
Concretely, in a linearized GNN with $L$ layers, the propagation of features is governed by the $L$-th power of the adjacency matrix, $\mathbf{A}^L$.
By performing a spectral decomposition, $\mathbf{A}^L$ can be expressed as $\sum_{k=1}^N \lambda_k^L \mathbf{v}_k \mathbf{v}_k^T$.
Notice that eigenvalues are squared in this expression, so the principal spectral component, which is $\mathbf{v}_k=c$,
is the primary driver of information diffusion.

\section{Performance evaluation}
\subsection{Datasets}\label{sec:datasets}
Three publicly available datasets were used in our experiments. Table \ref{tab:dataset_stats} shows relevant statistics, including the number of sensor nodes, data points, and the percentage of anomalies.
\par\vspace{0.5em}
\noindent\textit{SWaT (Secure Water Treatment)}~\cite{mathur2016swat} is a dataset collected from a testbed designed to replicate a six-stage water treatment process in Singapore. It serves as a representative cyber-physical system, integrating 25 sensors, 26 actuators, and programmable logic controllers to monitor and manage the production of clean water. The dataset contains 11 days of sensor readings, including 7 days of normal operations and 4 days involving a range of attack scenarios.
\par\vspace{0.5em}
\noindent\textit{WADI (Water Distribution)}~\cite{WADI2017} extends the SWaT setup by incorporating a network of distribution pipelines. It includes 16 days of operational data: 14 days under normal conditions and 2 days with injected physical attacks. 
\par\vspace{0.5em}
\noindent\textit{SJVAir}~\cite{SJVAir2017} contains PM2.5 measurements (\( \mu\text{g/m}^3 \)) collected at two-minute intervals from air quality sensors deployed across the city of Fresno, California.
The collected data is pre-processed by aggregating the two-minute readings into hourly averages. Sensors with more than one hour of consecutive missing data were excluded, resulting in a total of 41 sensors, each contributing 2,928 hourly measurements. Additionally, Fresno's historical wind data, comprising hourly wind speed and direction, are obtained from Visual Crossing~\cite{Visualcrossing2024} for the same time period. This meteorological data is aligned with the sampling frequency of the PM2.5 dataset to support multimodal analysis. 
We injected synthetically generated anomalies into the test data to simulate short bursts of random noise in sensor readings. These perturbations are sampled from a zero-mean Gaussian distribution, \(\mathcal{N}(0, \zeta^2 I_{L \times L})\), where \(\zeta = 10\) determines the intensity of the noise, and \(L\) is the anomaly duration drawn from the Poisson distribution with mean \( 7\). 
The identity matrix \(I_{L \times L}\) ensures that the noise is uncorrelated across the \(L\) time steps.

\begin{table}[t!]
\centering
\caption{Statistics of the three datasets used in experiments}
\label{tab:dataset_stats}
\begin{tabular}{lcccc}
\toprule
\textbf{Dataset} & \textbf{\# Nodes} & \textbf{\# Training} & \textbf{\# Test} & \textbf{Anomalies} \\
\midrule
SJVAir & 41  & 98,381  & 35,663  & 25\% \\
SWaT & 51  & 47,515  & 44,986  & 11.97\% \\
WADI & 127 & 118,795 & 17,275  & 5.99\%  \\

\bottomrule
\end{tabular}
\end{table}
\subsection{Evaluation Metrics}
We assess anomaly detection performance using F1 Score and Area Under the Precision-Recall Curve (AUC-PR) as metrics. Additionally, we use the Fractional Target Accuracy (FTA) used in prior work~\cite{zugner2018adversarial}. We define these metrics below.

Given the class imbalance typical in anomaly detection tasks, we report F1 Score which gives a balanced measure of the model's accuracy considering both precision and recall. 
Moreover, to evaluate the trade-off between precision and recall, we calculate and report AUC-PR by adjusting the decision threshold ($\lambda$). 
In our setting, $\lambda$ is chosen based on the maximum validation error on normal data and therefore varies across datasets.
Unlike AUC-ROC, which evaluates performance over both the positive and negative classes, AUC-PR emphasizes performance on the positive class. 
This makes AUC-PR more informative when anomalies are rare. A higher AUC-PR indicates that the model maintains high precision while achieving high recall, reflecting strong anomaly detection performance.

To evaluate the robustness of the anomaly detection model under adversarial attack, we report the FTA metric which measures the proportion of target nodes whose anomaly classification remains correct after perturbation, relative to the ground truth. For each target node, predictions are made at each time step using a sliding window of size $w$, and the node is considered correctly classified if its binary anomaly decision (based on a predefined threshold $\lambda$) matches the ground-truth label across the evaluated time steps.
Note that FTA decreases as the attack success rate increases.

In our implementation, the anomaly detection threshold ($\lambda$) is defined as the maximum error score observed on the validation set. These thresholds are specifically used for computing F1 score and FTA metrics. For the GDN model, the thresholds are 22.22 (SJVAir), 26.42 (SWaT) and 26.67 (WADI). In comparison, for the TopoGDN model, the thresholds are 15.19 (SJVAir), 20.90 (SWaT) and 19.33 (WADI). To evaluate the attack success rate, PGD is performed with a step size of $\alpha {=} 0.01$ for $t {=} 10$ iterations, to generate adversarial perturbations for assessing the model's robustness.

\begin{table*}[h]
\centering
\small
\caption{Anomaly detection performance in terms of F1 Score, AUC-PR and FTA for the GDN model under different adversarial attack strategies on SJVAir, SWaT, and WADI datasets.}
\begin{tabular}{lccc|ccc|ccc}
\toprule
\multirow{2}{*}{\textbf{Method}} 
& \multicolumn{3}{c|}{\textbf{SJVAir}} 
& \multicolumn{3}{c|}{\textbf{SWaT}} 
& \multicolumn{3}{c}{\textbf{WADI}} \\
\cmidrule(lr){2-4} \cmidrule(lr){5-7} \cmidrule(lr){8-10}
& \textbf{F1} & \textbf{AUC-PR} & \textbf{FTA} 
& \textbf{F1} & \textbf{AUC-PR} & \textbf{FTA} 
& \textbf{F1} & \textbf{AUC-PR} & \textbf{FTA} \\
\midrule
  No attack           & 0.9794 & 0.8250 & 0.8566    & 0.8521 & 0.9136 & 0.8364   & 0.7896 & 0.7848 & 0.9015    \\
    Random           & 0.3654 & 0.7243 & 0.7490    & 0.5737 & 0.8744 & 0.7261   & 0.5413 & 0.6998 & 0.8815       \\
Nettack          & 0.3526 & 0.6787 & 0.6557    & 0.5442 & 0.8338 & 0.6111   & 0.5303 & 0.5833 & 0.7989       \\
\textit{Ablation:} PGD + Heuristics      & 0.3461 & 0.5797 & 0.6141    & 0.4108 & 0.7176 & 0.5517   & 0.5234 & 0.4704 & 0.7543       \\
\textit{Ablation:} Nettack + GAFExplainer     & 0.2899 & 0.5497 & 0.3854    & 0.4034 & 0.5560 & 0.4425   & 0.5134 & 0.4667 & 0.7279       \\
\midrule
BETA (Budgeted)      & \textbf{0.2786} & \textbf{0.4392} & \textbf{0.3032} 
                & \textbf{0.3757} & \textbf{0.5060}  & \textbf{0.4330} 
                & \textbf{0.4531}   & \textbf{0.4110}   & \textbf{0.6836} \\
\bottomrule
\end{tabular}
\label{tab:gdn_anomaly_detection}
\end{table*}

\begin{table*}[h]
\centering
\small
\caption{Anomaly detection performance in terms of F1 Score, AUC-PR and FTA for the TopoGDN model under different adversarial attack strategies on SJVAir, SWaT, and WADI datasets.}
\begin{tabular}{lccc|ccc|ccc}
\toprule
\multirow{2}{*}{\textbf{Method}} 
& \multicolumn{3}{c|}{\textbf{SJVAir}} 
& \multicolumn{3}{c|}{\textbf{SWaT}} 
& \multicolumn{3}{c}{\textbf{WADI}} \\
\cmidrule(lr){2-4} \cmidrule(lr){5-7} \cmidrule(lr){8-10}
& \textbf{F1} & \textbf{AUC-PR} & \textbf{FTA} 
& \textbf{F1} & \textbf{AUC-PR} & \textbf{FTA} 
& \textbf{F1} & \textbf{AUC-PR} & \textbf{FTA} \\
\midrule
 No attack           & 0.8480 & 0.9204 & 0.8752  & 0.8774 & 0.8903 & 0.8931   & 0.9048 & 0.9192 & 0.9854   \\
Random           & 0.7880 & 0.7428 & 0.7881  & 0.8028 & 0.8840 & 0.7993   & 0.8025 & 0.8342 & 0.9650     \\
Nettack           & 0.7219 & 0.7033 & 0.6852  & 0.7206 & 0.8463 & 0.7835   & 0.7727 & 0.8138 & 0.9362     \\
\textit{Ablation:} PGD + Heuristics       & 0.6353 & 0.5903 & 0.6564  & 0.6764 & 0.8300 & 0.7203   & 0.7211 & 0.7806 & 0.9136     \\
\textit{Ablation:} Nettack + GAFExplainer      & 0.5829 & 0.5576 & 0.5802  & 0.5659 & 0.7147 & 0.6533   & 0.6154 & 0.6365 & 0.7963     \\
\midrule
BETA (Budgeted)      & \textbf{0.5532} & \textbf{0.4566} & \textbf{0.4547} 
                & \textbf{0.4390} & \textbf{0.5893}  & \textbf{0.6418} 
                & \textbf{0.5557}   & \textbf{0.6153}   & \textbf{0.7387} \\
\bottomrule
\end{tabular}
\label{tab:topogdn_anomaly_detection}
\end{table*}

\subsection{Baselines}
We use the same target node in all attacks described below.

\subsubsection{Random attack}
In the random attack, $B$ influencer nodes are randomly selected from the graph (excluding the target) and their features are perturbed.
Specifically, each feature of an influencer node is replaced with a new value drawn from a uniform distribution \( \mathcal{U}(\min_j, \max_j) \), where \( \min_j \) and \( \max_j \) denote the minimum and maximum values of feature \( j \) across the dataset. 

\subsubsection{Nettack~\cite{zugner2018adversarial}}
The original graph \( G = (V, \mathcal{E}) \) is transformed into an adversarial version \( G' = (V, \mathcal{E}') \) with a modified feature matrix \( X'^t \), such that a designated \textit{target node} \( u\) is misclassified by the GNN model. The attack introduces minimal yet effective perturbations by modifying a small subset of \textit{attacker nodes}. 
The attacker nodes are chosen based on how much each neighbor of the target node \( u \) influences its classification. To do this, Nettack approximates the effect of modifying each candidate edge by computing the output logits using the second-order adjacency matrix and the node feature weights. For each candidate edge, it then calculates a score as the difference between the logits of the true class of the target node and the maximum logit of all incorrect classes. To ensure the true class does not affect the max operation, its logit is suppressed by subtracting a large constant. Edges that receive lower scores are those that reduce the model's confidence in the correct label, making them stronger candidates for attack. By ranking the neighbors with this score, Nettack selects the top-B influencer nodes to manipulate during the attack.

Nettack performs two types of modifications: (1) \textit{}{Feature Manipulation}, which alters the node feature matrix \( X^{t} \) to deceive the model's decision process, and (2) \textit{Structural Modification}, which updates the edge set by adding or removing edges, thereby affecting the neighborhood context and information propagation around the target node. 
For a fair comparison with BETA which is a feature-based evasion attack, we implement a variant of Nettack that performs feature manipulation only, without altering the graph structure.

\subsubsection{Nettack + GAFExplainer} We use the same strategy for selecting candidate influencer nodes and pruning that set as BETA (Section~\ref{sec:gafexplainer} and~\ref{sec:centrality}).
However, instead of using PGD, Nettack's feature perturbation strategy is applied to these influencer nodes to mislead the model's prediction for the target node. 
This baseline serves as an ablation study to evaluate the effectiveness of constrained PGD compared with Nettack's feature modification approach.

\subsubsection{PGD + Heuristics} This baseline combines our constrained PGD attack with the influencer node selection strategy from Nettack. Once these nodes are selected, PGD attack is applied to their features to craft adversarial examples that stay within a perturbation budget and preserve stealth, while maximizing the chance of misclassifying the target node.
This baseline serves as an ablation study to evaluate the effectiveness of our proposed node selection strategy compared with the heuristic selection approach used in Nettack.

\subsubsection{BETA (Unbudgeted)} In Section~\ref{sec:sensitivity-budget}, we apply BETA in a non-budgeted setting, assuming all nodes in the graph, except the target node, are influencer nodes. This setup helps obtain a bound on the performance of our attack when not constrained by a budget.

\subsection{Experimental Setup}

We implemented GDN and TopoGDN using PyTorch~\cite{paszke2019pytorch}, following the authors' implementation of GDN\footnote{\url{https://github.com/d-ailin/GDN}} and TopoGDN\footnote{\url{https://github.com/ljj-cyber/TopoGDN}}. The Nettack implementation from its public repository was adapted to remove the structural modification component, which is not applicable for feature-based attacks\footnote{\url{https://github.com/danielzuegner/nettack}}. We used the GAFExplainer implementation provided by the authors\footnote{\url{https://github.com/wyhi/GAFExplainer}}. We followed the preprocessing strategy adopted in TopoGDN~\cite{liu2024multivariate}, applying Min-Max normalization to scale all sensor readings to the 
[0,1] range. 
For PGD, we performed independent trials, each initialized from a random point within the $\ell_\infty$-ball around the original input. We started from 5 randomly selected feature matrices, i.e. $\tilde{\mathbf{X}}$.
Perturbations with an $\ell_\infty$ norm smaller than $\epsilon=0.1$ were deemed imperceptible. 
We segmented the time series using a sliding window of size 100 with a stride of 10. The model was trained using the Adam optimizer, with a learning rate of 0.001 and a batch size of 32, on an NVIDIA RTX 2080 TI GPU.

\section{RESULTS AND ANALYSIS}
We evaluate the performance of several attack strategies across 3 datasets, SJVAir, SWaT, and WADI, using F1 Score, AUC-PR and FTA.

\subsection{Attack Success Evaluation}
Table~\ref{tab:gdn_anomaly_detection} reports GDN's anomaly detection performance under different attack scenarios with a perturbation budget of $B{=}5$ sensors. 
In the absence of adversarial perturbations (i.e., the no-attack scenario), GDN exhibits strong anomaly detection performance, with high F1, AUC-PR, and FTA scores in all datasets. 
However, GDN's F1 score decreases sharply, by an average of 50.45\% across the three datasets, when evaluated on data perturbed by BETA.
Among the baseline attacks, Random results in the smallest drop in anomaly detection performance, particularly with respect to AUC-PR and FTA.
Nettack outperforms Random but remains less effective than BETA and its two variants, particularly in terms of AUC-PR and FTA. Among the three datasets, 
WADI exhibits the smallest degradation in GDN's anomaly detection performance under all attack scenarios.

The ablation study demonstrates the contribution of each major component of BETA, namely constrained PGD for feature perturbation and GAFExplainer combined with eigenvector-centrality-based pruning for influencer node selection. PGD + Heuristics improves upon Nettack by applying gradient-based perturbations to node features. In contrast, Nettack + GAFExplainer leverages GNN explanations to guide influencer node selection, yielding stronger attack performance than Random, Nettack, and even PGD + Heuristics. 
Across SJVAir, SWaT, and WADI, the results indicate that the improvement in attack success is primarily attributable to the influencer node selection strategy employed by BETA, whereas the choice of perturbation method plays a comparatively smaller role.

Table~\ref{tab:topogdn_anomaly_detection} shows how adversarial attacks affect TopoGDN’s anomaly detection performance when the perturbation budget is $B{=}5$ sensors. 
In the absence of adversarial perturbations, TopoGDN consistently achieves high performance, with F1, AUC-PR, and FTA scores above 84\% on all three datasets. Under adversarial attacks, BETA achieves the highest attack success, and the ranking of the baseline methods is consistent with that observed for GDN. Notably, TopoGDN's F1 score decreases by an average of 36.07\% across the three datasets when evaluated on data perturbed by BETA. Overall, although TopoGDN is more robust to adversarial attacks than GDN, BETA still induces a significant drop in anomaly detection performance.

\subsection{Sensitivity to Attack Budget}\label{sec:sensitivity-budget}

\begin{figure*}[t!]
\centering
\includegraphics[width=.9\linewidth]{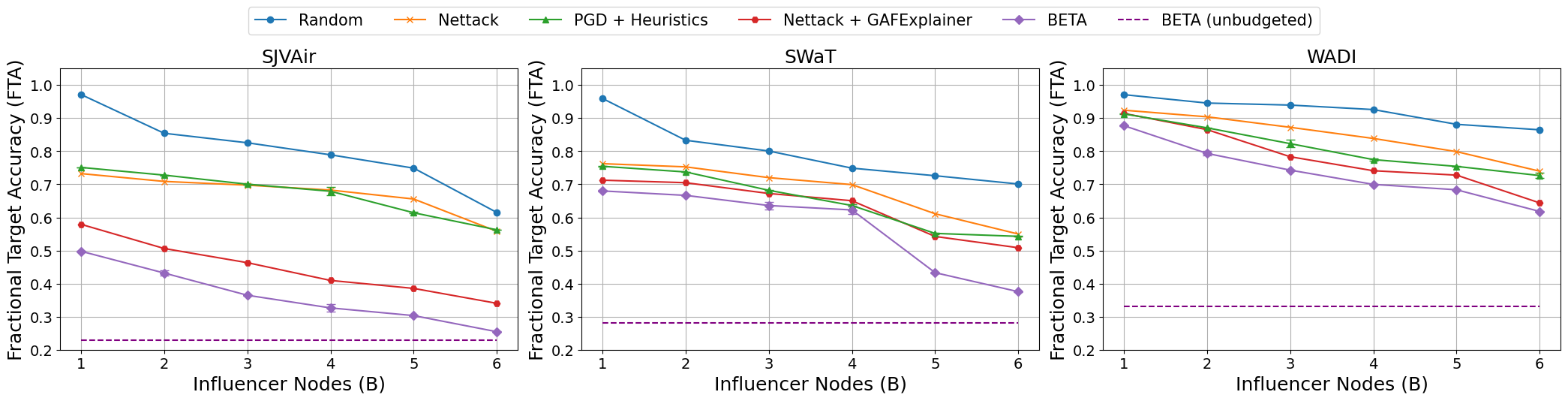}
\vspace{-2mm}
\caption{FTA comparison of attack methods with varying influencer nodes on the GDN model using SJVAir, SWaT and WADI dataset. Error bars show standard
deviation across 5 runs.}
\label{fig:GDN}
\end{figure*}

We compare the effectiveness of the attack strategies under varying perturbation budgets using the FTA metric. Specifically, we evaluate how the attack performance changes as the perturbation budget increases from one to six influencer nodes. We first consider the effectiveness of the attack strategies against GDN. As depicted in Figure~\ref{fig:GDN}, a clear trend emerges across all three datasets: increasing the perturbation budget consistently reduces FTA, indicating more successful attacks. BETA consistently achieves the lowest FTA across all budget levels, underscoring the benefit of combining our influencer node selection strategy with constrained PGD.

Next, we examine the performance gap between BETA, the strongest baseline (Nettack + GAFExplainer), and the unbudgeted variant of BETA across the three datasets. In SJVAir, FTA under BETA drops from 0.4978 at $B{=}1$ to 0.2544 at $B{=}6$, whereas Nettack + GAFExplainer reduces FTA from 0.5792 to 0.3403. Moreover, the FTA achieved by BETA at $B{=}6$ is already close to that of the unbudgeted variant (0.2298), suggesting that only a small number of carefully selected influencer nodes are sufficient to launch a powerful attack against GDN.

In SWaT, FTA under BETA decreases from 0.6801 at $B{=}1$ to 0.3755 at $B{=}6$, compared with a decrease from 0.7126 to 0.5084 for Nettack + GAFExplainer. The unbudgeted variant of BETA further reduces FTA to 0.2816. In WADI, which is the largest dataset and has the lowest anomaly ratio, FTA under BETA decreases from 0.8775 at $B{=}1$ to 0.6181 at $B{=}6$, while Nettack + GAFExplainer reduces FTA from 0.9145 to 0.6439. Although the performance differences among the attack methods are smaller on WADI, BETA consistently achieves the lowest FTA across all budget levels.

\begin{figure*}[t!]
\centering
\includegraphics[width=.9\linewidth]{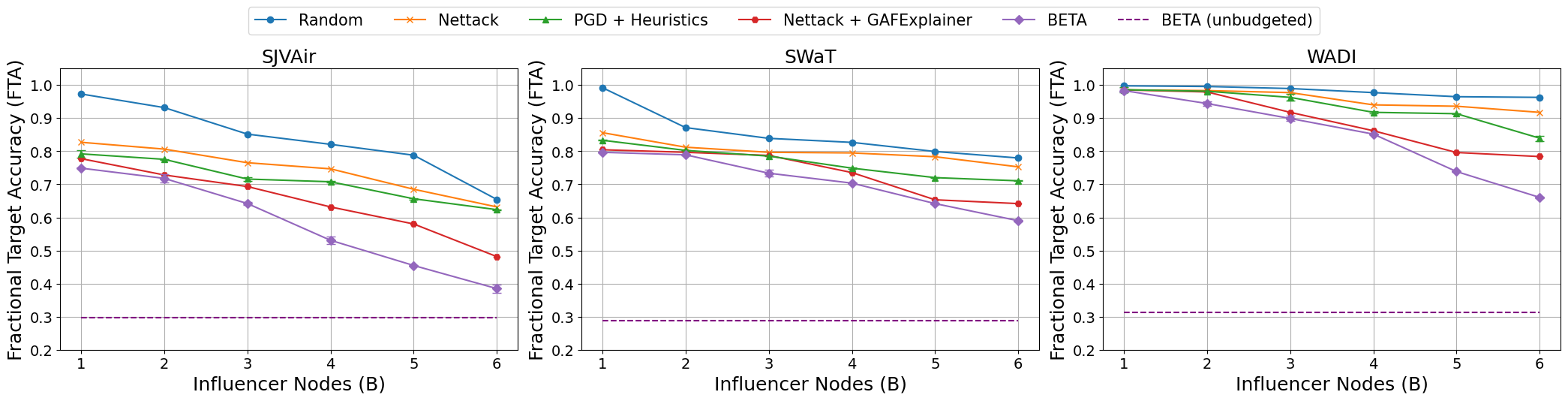}
\vspace{-2mm}
\caption{FTA comparison of attack methods with varying influencer nodes on the TopoGDN model using SJVAir, SWaT and WADI dataset. Error bars show  standard
deviation across 5 runs.}
\label{fig:TopoGDN}
\end{figure*}

Figure~\ref{fig:TopoGDN} compares the performance of these attack strategies against TopoGDN under varying perturbation budgets. Similar to GDN, increasing the perturbation budget consistently lowers FTA, although TopoGDN is generally more robust to feature-based evasion attacks. Nevertheless, BETA again achieves the lowest FTA across all budget levels.
We note that the relatively large gap between BETA and its unbudgeted variant suggests that perturbing more influencer nodes may be necessary to effectively fool TopoGDN through an indirect evasion attack.

\subsection{Centrality Measures for Pruning Influencer Nodes}
\noindent Table~\ref{tab:fta_centrality} compares the impact of various centrality-based node selection strategies that can be used to prune the set of influencer nodes identified by GAFExplainer and evaluates their impact on the effectiveness of BETA under perturbation budgets ranging from $B{=}3$ to $B{=}6$ on the SJVAir dataset.
Specifically, we rank nodes in $\mathcal{V}_S$ using six well-established centrality measures~\cite{newman2010networks}: eigenvector centrality, node degree, closeness centrality, betweenness centrality, clustering coefficient, and average neighbor degree. 
Across all perturbation budgets, selecting influencer nodes based on eigenvector centrality consistently achieves the lowest FTA, highlighting its superior ability to identify nodes that most effectively propagate the injected perturbations through the graph. For instance, at $B{=}6$, eigenvector-centrality-based selection achieves an FTA of 0.2011, outperforming all alternative centrality measures.

\begin{table}[t!]
\centering
\small
\caption{Impact of centrality metrics on the attack's effectiveness (with respect to FTA) against GDN on the SJVAir dataset.}
\label{tab:fta_centrality}
\begin{tabular}{lcccc}
\toprule
Centrality measure & $B=3$ & $B=4$ & $B=5$ & $B=6$ \\
\midrule
\textbf{Eigenvector}          & \textbf{0.3029} & \textbf{0.3011} & \textbf{0.2630} & \textbf{0.2011} \\
Degree               & 0.3962 & 0.3043 & 0.2836 & 0.2398 \\
Clustering  Coefficient         & 0.3998 & 0.3410 & 0.3005 & 0.2167 \\
Betweenness          & 0.3634 & 0.3235 & 0.3013 & 0.2521 \\
Closeness            & 0.3469 & 0.3125 & 0.2761 & 0.2504 \\
Avg. Neighbor Degree  & 0.3635 & 0.3675 & 0.2701 & 0.2048 \\
\bottomrule
\end{tabular}
\end{table}
\subsection{Impact of Combining Explanation-Based Guidance and Centrality-Based Node Selection}
\begin{figure}[t!]
\centering
\includegraphics[width=0.8\linewidth]{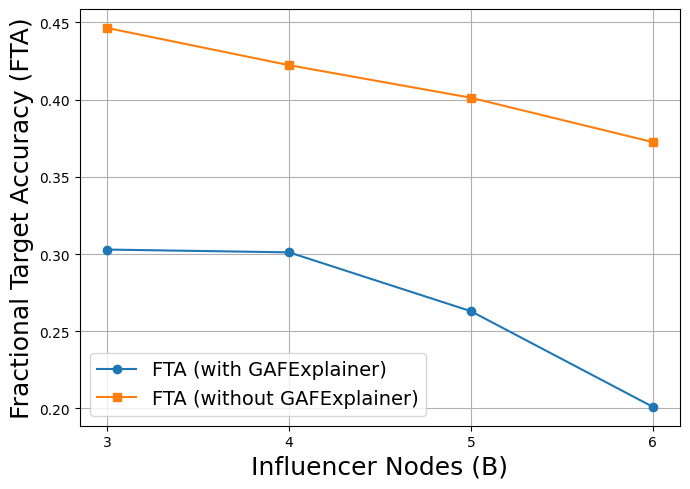}
\caption{Comparing the impact of combining GAFExplainer and eigenvector centrality or using only eigenvector centrality for selecting the influencer nodes on the GDN model's FTA on the SJVAir dataset}
\label{fig:Centrality}
\end{figure}

\noindent We investigate the effectiveness of combining GAFExplainer with eigenvector centrality for influencer node selection by comparing it with a strategy that relies solely on eigenvector centrality. For this study, we evaluate GDN on the SJVAir dataset using these two node selection strategies within BETA. 
As shown in Figure~\ref{fig:Centrality}, using only eigenvector centrality yields higher FTA values, ranging from 0.4464 for $B{=}3$ to 0.3725 for $B{=}6$, indicating less effective attacks. In contrast, combining GAFExplainer with eigenvector centrality consistently lowers FTA, from 0.3029 for $B{=}3$ to 0.2011 for $B{=}6$, indicating more effective attacks. These results underscore the importance of identifying nodes that strongly influence the target node's prediction and refining the selection, if necessary, by considering the multi-hop propagation of adversarial perturbations throughout the graph.

\section{Conclusion and Limitations}
We introduced BETA (Budgeted Explainability-guided Targeted Attack), a novel budgeted evasion attack against GNNs deployed for anomaly detection in sensor networks. 
Unlike traditional unconstrained graph adversarial attacks, BETA operates under strict, physically realistic constraints where the primary target sensor is heavily safeguarded against direct cyber or physical compromise, and the adversary must operate under a strict budget.
To solve the intractable optimal influencer node selection and perturbation generation problem, BETA repurposes post-hoc GNN explanations to identify structural vulnerability in the GNN. To meet the strict budget constraint, it employs a spectral pruning scheme based on eigenvector centrality.
Bounded continuous noise is then injected into the selected nodes using PGD.

Extensive empirical evaluations across three real-world testbeds against representative GNN-based anomaly detectors demonstrate the efficacy of our attack and the vulnerability of current GNN-based detectors. BETA consistently outperformed baseline strategies, reducing detector F1-scores by up to 39.16\% through entirely indirect, imperceptible feature perturbations. Crucially, these findings expose a fundamental security blind spot in graph-based anomaly detection: safeguarding the critical target sensor(s) is entirely ineffective if peripheral sensors that emit correlated data with the target sensor remain unsecured against adversarial manipulation.

\textcolor{black}{Our work has few limitations. First, 
the attacker is assumed to be capable of compromising any sensor apart from the target sensor; however, real deployments may exhibit heterogeneity where every sensor cannot be compromised with the same cost. In such settings, the node selection strategy should weight the cost of compromising each sensor against its influence. 
In the limiting case, the cost of compromising some sensor may be infinity, as a result they must be removed if they appear in the set of candidate influencer nodes, which may reduce attack effectiveness. 
Second, using GAFExplainer to reduce the search space for the attacker increases the computational overhead of the attack.}
Future work will focus on (a) developing robust defenses against such attacks, including conformal anomaly detection, attribution and perturbation-score-based techniques; and (b) evaluating BETA against graph-based anomaly detection models with a dynamically updated structure.

\bibliographystyle{ACM-Reference-Format}
\bibliography{ref}

\end{document}